\documentclass[sigconf,nonacm]{acmart} 
\usepackage{cleveref}
\usepackage{amsmath}
\usepackage{newtxtext}
\usepackage{newtxmath}

\AtBeginDocument{%
  }

\begin{document}

\title{Unleashing Diverse Thinking Modes in LLMs through Multi-Agent Collaboration}

\author{\bf{Zhixuan He \quad Yue Feng}}
\email{hezhixuan1997@gmail.com}
\email{y.feng.6@bham.ac.uk}
\affiliation{%
  \institution{University of Birmingham, United Kingdom}
  \country{}
}



\begin{abstract}
Large Language Models (LLMs) demonstrate strong performance but often lack interpretable reasoning. This paper introduces the Multi-Agent Collaboration Framework for Diverse Thinking Modes (DiMo), which enhances both performance and interpretability by simulating a structured debate among four specialized LLM agents. Each agent embodies a distinct reasoning paradigm, allowing the framework to collaboratively explore diverse cognitive approaches. Through iterative debate, agents challenge and refine initial responses, yielding more robust conclusions and an explicit, auditable reasoning chain. Across six benchmarks and under a unified open-source setup, DiMo improves accuracy over widely used single-model and debate baselines, with the largest gains on math. We position DiMo as a semantics‑aware, Web‑native multi‑agent framework: it models human–machine intelligence with LLM agents that produce semantically typed, URL‑annotated evidence chains for explanations and user‑friendly interactions. Although our experiments use standard reasoning benchmarks, the framework is designed to be instantiated over Web corpora and knowledge graphs, combining retrieval‑augmented reasoning with structured justifications that downstream systems can inspect and reuse. 
\end{abstract}

\keywords{LLM agent, Multi-Agent Debate}

\received{20 February 2007}
\received[revised]{12 March 2009}
\received[accepted]{5 June 2009}

\maketitle
\settopmatter{printacmref=false}
\fancyhead{} 
\pagestyle{plain}

\section{Introduction}
In recent years, Large Language Models (LLMs), through continuous iteration and evolution aided by the scaling laws and pretraining techniques, have demonstrated remarkable performance across different web tasks \cite{llama3modelcard, liu2024deepseek, team2025kimi}.
While individual LLMs have demonstrated impressive capabilities in various tasks, their potential limitations in addressing multifaceted problems have led researchers to explore collaborative and adversarial interactions between multiple model instances. Multi-agent debate systems, where multiple LLMs engage in structured discourse, represent a promising direction for enhancing reasoning capabilities and reducing individual model biases \cite{guo2024large}. This approach draws inspiration from human deliberative processes, where perspectives and critical discourse often lead to more robust conclusions. Such systems have shown particular promise in tasks requiring complex reasoning, fact-checking, and the evaluation of competing hypotheses \cite{wang2023apollo}.\\

On the other hand, the development of LLMs also faces several significant challenges, notably the interpretability of LLMs. Due to the black-box nature of LLMs, their problem-solving mechanisms cannot be explicitly observed \cite{zhao2024explainability}. For instance, when handling reasoning tasks, it remains unclear whether the LLMs arrive at correct answers through genuine reasoning processes or merely reference knowledge acquired during the pretraining stage. Investigating how to address the interpretability of LLMs not only helps reveal the internal reasoning logic of the LLMs but also provides crucial insights for improving model design and reducing errors.\\
\begin{figure}[t]
\begin{center}
\centerline{\includegraphics[width=\columnwidth]{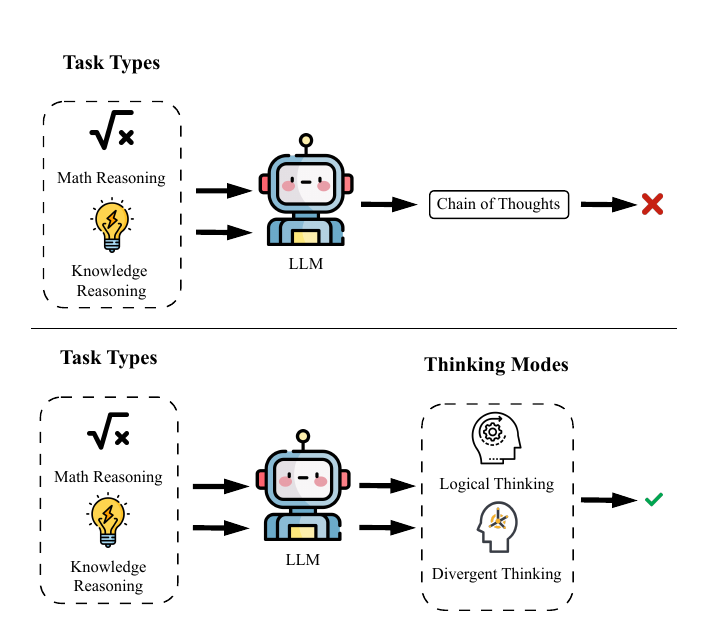}}
\caption{LLMs prefer to use different thinking modes to solve different reasoning tasks and lead to higher accuracy.}
\vspace{-8 mm}
\label{fig:problem}
\end{center}
\end{figure}

In this work, we focus on LLMs' reasoning capabilities across different reasoning tasks and their interpretability during the reasoning process. And we address a Web and semantics problem: modeling human–machine intelligence with LLM agents that output machine‑interpretable, provenance‑aware explanations. We propose an LLM-based multi-agent debate framework called Multi-Agent Collaboration Framework for Diverse Thinking Modes(DiMo) to enhance both the LLMs' reasoning capabilities among different reasoning tasks and their interpretability in reasoning processes. In this framework, we establish four LLM agents with different roles to engage in collaboration and debate among different reasoning tasks. Through iterative deliberation, these agents identify and rectify issues, ultimately outputting correct solutions that enhance LLMs' reasoning capabilities. We instantiate DiMo as a Web‑native, semantics‑aware pipeline—retrieve passages, link entities/relations, cross‑check multi‑document evidence, debate, and verify claims—yielding URL‑annotated, typed reasoning paths. Additionally, the framework explicitly generates reasoning paths, which improves LLM interpretability.\\

\textbf{Motivation.} Large language models excel across benchmarks, yet their reasoning remains brittle and opaque: \textbf{single‑model prompting can swing between correct solutions and confident mistakes, and intermediate steps are rarely auditable. Debate‑style multi‑agent systems improve robustness by letting agents critique and revise one another, but they often lack operational definitions of the “modes” of thinking being exercised, blur process transparency with mechanistic interpretability, and are hard to compare fairly across compute budgets.} Our goal is to provide a reproducible, open‑source protocol that makes these design choices explicit. DiMo constrains collaboration into two operational modes—divergent (parallel hypothesis/knowledge proposal) and logical (step‑wise verification with localized refinement)—so we can ask not only “does debate help?” but “which protocol helps which task under a fixed budget?” \textbf{This lens lets us study protocol–task affinity while improving answer accuracy and exposing auditable traces, without making claims about the base models’ internal mechanisms.}

We conduct experiments on two types of reasoning tasks: commonsense and knowledge-based reasoning and mathematical reasoning. We select recent and challenging datasets within these domains to ensure both the contemporary relevance and complexity of our experiments. Commonsense reasoning requires LLMs to analyze and identify relationships among knowledge entities within the questions, while mathematical reasoning demands the application of logical reasoning capabilities to compute solutions. Our experiments reveal that LLMs benefit from different thinking modes when addressing various reasoning tasks. For example, when handling mathematical reasoning questions, LLMs perform better in the rigorous logical thinking mode. The experimental results demonstrate that our DiMo approach achieves enhanced reasoning performance among the majority of datasets compared to baseline methods. And the framework outputs explicit reasoning paths, revealing how LLMs reach their conclusions.\\
\begin{figure}[t]
\begin{center}
\centerline{\includegraphics[width=\columnwidth]{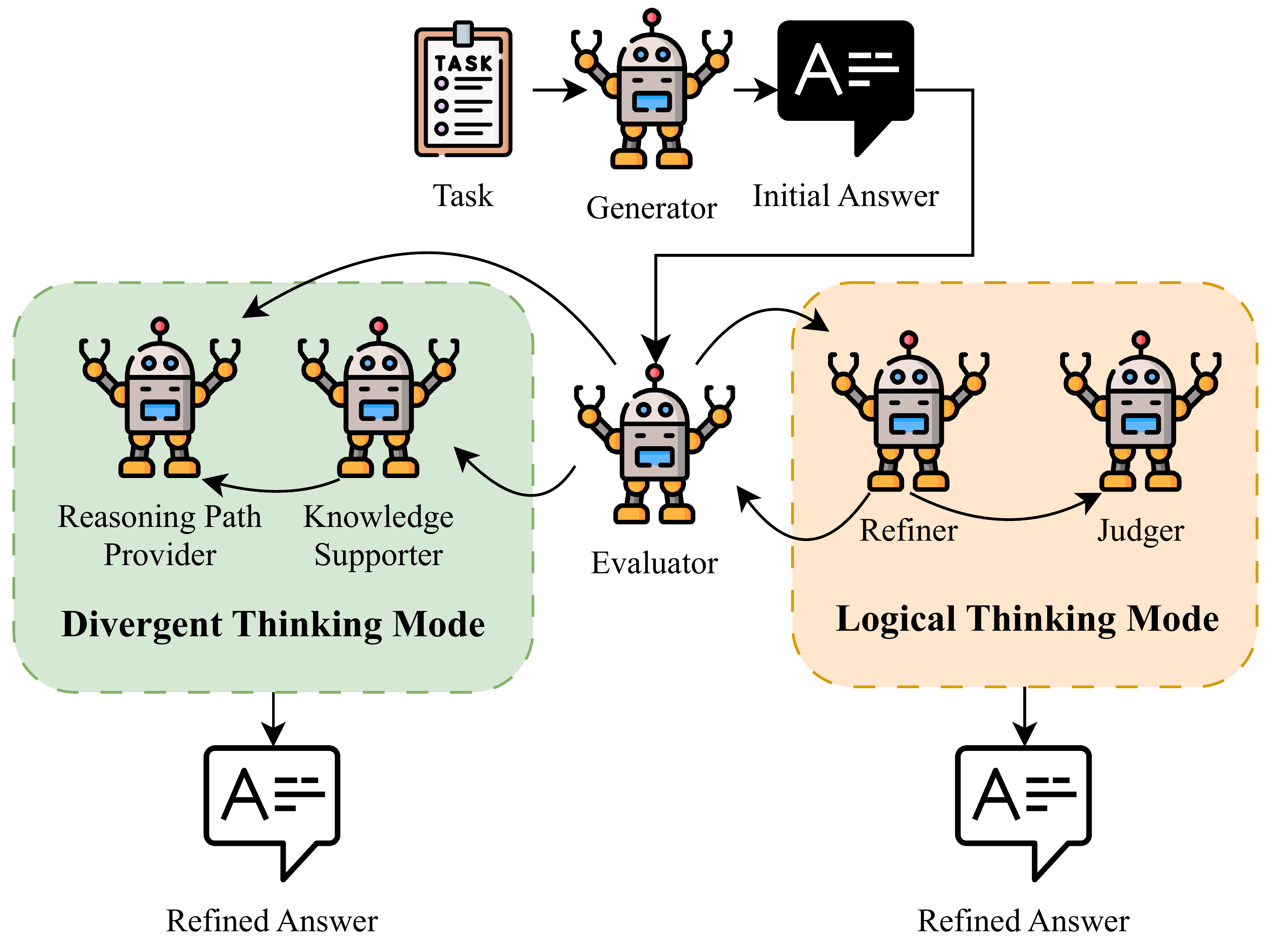}}
\caption{An Overview of Multi-Agent Collaboration Framework for Diverse Thinking Modes (DiMo). Role‑specialized agents debate and refine solutions, yielding auditable intermediate reasoning traces.}
\vspace{-8 mm}
\label{fig:overview}
\end{center}
\end{figure}

Our contributions include: 1)We formalize a Multi‑Agent Collaboration Framework (DiMo) that instantiates two human‑inspired yet strictly operational “thinking modes”—divergent and logical—via explicit roles and interaction rules \cref{fig:problem}. The modes constrain which intermediate states can be proposed and how they are critiqued, giving an engineering‑level abstraction; neuroscience citations are used as analogy‑level motivation, not mechanistic evidence. 2) DiMo makes the reasoning process auditable: it externalizes hypotheses, supportive knowledge, and step‑wise refinements so that other agents (and humans) can inspect, challenge, and revise them. We therefore scope “interpretability” as process transparency rather than mechanistic interpretability, and we highlight how our protocol differs from prior debate systems while retaining alignment with human‑style problem solving. 3) Under a unified token budget and a reproducible open‑source setup, DiMo improves answer accuracy across six benchmarks—with the largest gains on math—and we analyze where it helps or not. We also provide factor analyses (debate rounds, initialization prompts) and a complexity/cost accounting that decomposes tokens and latency to inform fair comparison and deployment. 4) Most importantly, we use DiMo as an analytic lens to characterize a protocol–task affinity: in our setting, commonsense tasks tend to benefit more from the divergent protocol, whereas math tasks benefit from the logical protocol. We present this as an observational regularity—conditional on models, prompts, judges, and budgets—rather than an intrinsic model preference, and we discuss learned routing as future directions.

\section{Related Work}

\textbf{Reasoning in Large Language Models.} 
Recent research has extensively explored the reasoning capabilities of Large Language Models (LLMs). Previous studies have demonstrated that LLMs possess sophisticated reasoning abilities but exhibit distinct limitations in different reasoning domains \cite{brown2020language,wei2022chain}.
Chain-of-Thought prompting has emerged as a pivotal advancement in enhancing LLMs' reasoning capabilities, enabling models to break down complex problems into intermediate steps \cite{wei2022chain,kojima2022large}. Building upon this foundation, subsequent research has introduced various prompting techniques, including Tree-of-Thoughts \cite{yao2024tree} and Self-Consistency \cite{wang2022self}, which further improve reasoning performance through structured thinking processes. Recent works have also explored post-training approaches for LLMs using reinforcement learning and long Chain-of-Thoughts. These methods have been demonstrated to enhance LLMs' performance on mathematical reasoning tasks \cite{deepseekai2025deepseekr1,qwq-32b-preview}.\\

\textbf{Self Improvement in Large Language Models.} 
Recent studies in Large Language Models have demonstrated their potential for self-reflection and self-refinement, enabling them to iteratively improve their outputs. Researches such as those by Madaan et al. introduce frameworks like Self-Refine, where LLMs generate initial outputs, critique their own responses, and refine them based on iterative feedback \cite{madaan2024self}.
Similarly, Renze \& Guven investigate the effects of self-reflection in LLM agents on problem-solving tasks, showing that LLM agents can significantly improve their performance by reflecting on their mistakes and generating guidance for future attempts \cite{renze2024self}. Their work highlights the importance of decomposing self-reflection into components such as explanations, instructions, and solutions, which contribute to performance gains across multiple LLMs and problem domains.
Additionally, Shinn et al. propose Reflexion, a framework that leverages verbal reinforcement to enable LLM agents to learn from trial-and-error through self-reflection \cite{shinn2024reflexion}. By maintaining an episodic memory of reflective feedback, Reflexion agents improve decision-making, reasoning, and programming tasks, achieving state-of-the-art results on leveraging meta-cognitive strategies in LLMs that improve performance on downstream tasks and foster more robust and interpretable AI.\\

\textbf{Multi-Agent Debate.}
In recent years, the exploration of Multi-Agent Debate System systems leveraging Large Language Models (LLMs) has garnered significant attention in the field of artificial intelligence. Researchers have increasingly focused on harnessing the capabilities of LLMs to simulate collaborative and adversarial interactions among multiple agents.
For instance, studies such as those by Xiong et al., Du et al., and Liang et al. have demonstrated that multi-agent debate frameworks can enhance decision-making accuracy and robustness by enabling agents to critique, refine, and consolidate diverse perspectives \cite{xiong2023examining,du2023improving,liang2023encouraging}. These systems often employ iterative dialogue mechanisms, where agents propose hypotheses, challenge each other's reasoning, and iteratively converge toward more refined solutions. This body of work underscores the potential of multi-agent debate systems to advance complex problem-solving tasks while highlighting critical areas of future research.\\

\textbf{Large Language Models' Interpretability.} 
Recent research has focused significantly on the interpretability of Large Language Models. Luo et al. propose a survey \cite{luo2024understanding} that examines LLM explainability methods and their applications in model editing and enhancement, bridging theory and practice to develop more transparent AI systems. Wang et al. present LSP (LLM-based Symbolic Programs)\cite{wang2024large}, integrating LLMs with symbolic rules to achieve both expressiveness and interpretability in predictive models with superior results. Creswell et al. introduce a Selection-Inference framework\cite{creswell2022selection} that improves large language models' logical reasoning capabilities while providing interpretable reasoning traces.

\section{Methodology}

We propose a LLM-based multiagent debate framework--Multi-Agent Collaboration Framework for Diverse Thinking Modes (DiMo). We provide an overview of our method in \cref{fig:overview}. Within this framework, we establish two distinct thinking modes to enable LLMs to effectively handle various types of reasoning tasks. We set an agent called Generator to receive input questions and generate initial answers. This framework is inspired by the observation that the human brain employs different thinking modes when confronting various reasoning problems. Osherson et al. have found that PET scans reveal deductive and probabilistic reasoning activate distinct brain regions, with probabilistic reasoning engaging left frontal areas and deductive reasoning activating right hemisphere occipital-parietal regions. \cite{osherson1998distinct} Castaeda et al. present a fMRI study\cite{castaneda2023probabilistic} that reveals that deductive and probabilistic reasoning activate distinct brain regions, confirming that they are fundamentally cognitive processes at the neural level. \\

\textbf{Definitions of Modes.} We define a "mode" as an interaction protocol that restricts what types of intermediate states an agent may produce and how other agents may critique or modify them. Divergent mode requires proposing alternative hypotheses, supportive knowledge snippets, and candidate reasoning paths. Logical mode enforces stepwise derivation with verification and refinement.\\

\textbf{Generator.} The Generator functions as a foundational component in the DiMo, serving as an intelligent intermediary between task input and subsequent processing layers. This module implements comprehensive parsing mechanisms to process incoming tasks and utilizes domain-specific heuristics to construct initial answers. Through its adaptive processing capabilities, the Generator produces initial answers.
Following the generation of the initial answers, the output is directed to the Evaluator for assessment.\\

\textbf{Evaluator.} The Evaluator functions as the system's comprehensive quality assessment module. Based on the types of tasks, the Evaluator employs different assessment strategies. For tasks that are dependent on factual knowledge, the Evaluator will identify logical deficiencies and knowledge gaps in the answers and generate structured evaluation reports. For the tasks that rely on logical reasoning processes, the Evaluator will implement step-by-step verification protocols to identify specific instances of logical inconsistencies and computational errors within the reasoning paths. The verification is characterized by its granular approach to error detection, focusing on both accuracy and logical coherence. Following the Evaluator's assessment, initial answers are channeled into different thinking modes according to the task type, where they undergo iterative refinement and enhancement through the multi-agent debate process.\\

We design distinct prompts for both the Generator and the Evaluator for different types of reasoning tasks. For commonsense reasoning tasks, we make the Generator output the initial answer directly, while for mathematical reasoning tasks, the Generator is required to output an initial answer with a complete step-by-step solution. We present the sample prompt designed for the Generator in the mathematical reasoning tasks:
    \textit{You are a mathematical problem solver focused on understanding and solving problems systematically. Follow these steps:1. Problem Understanding: a. Extract the core question (what exactly needs to be calculated) b. Identify all explicit and implicit conditions c. List all given values with their units d. Describe key relationships between values. 2. Solution Planning: a. Break down the problem into logical steps b. Identify the sequence of calculations needed c. Note any special conditions or edge cases d. Plan how to validate intermediate results. 3. Detailed Solution: a. Write clear, step-by-step calculations b. Include units in every calculation c. Show all work explicitly d. Validate each intermediate result. 4. Answer Validation: a. Verify the final calculation b. Check units in the final answer c. Confirm the answer makes logical sense d. Cross-check against the original question.}
We can construct the Generator with a specific formula as follows:
\begin{equation}
    A=Generator (Q),
\end{equation}
Where $A$ represents the initial answer, $G$ represents the Generator and $Q$ represents the input question. Previous research indicates that LLMs employ distinct processing mechanisms when addressing some tasks that are heavily dependent on factual knowledge, such as commonsense reasoning tasks, versus some tasks that require deriving solutions through logical reasoning processes, such as mathematical reasoning tasks. \cite{brown2020language} The empirical evidence suggests a fundamental difference in how these models approach and resolve these two categories of cognitive challenges. Where commonsense reasoning often relies on broad contextual understanding and implicit knowledge integration, mathematical reasoning typically involves more structured, rule-based processing paths. To further analyze whether LLMs exhibit preferential thinking modes for different tasks, we design two distinct thinking modes.
\begin{figure}[t]
\begin{center}
\centerline{\includegraphics[width=\columnwidth]{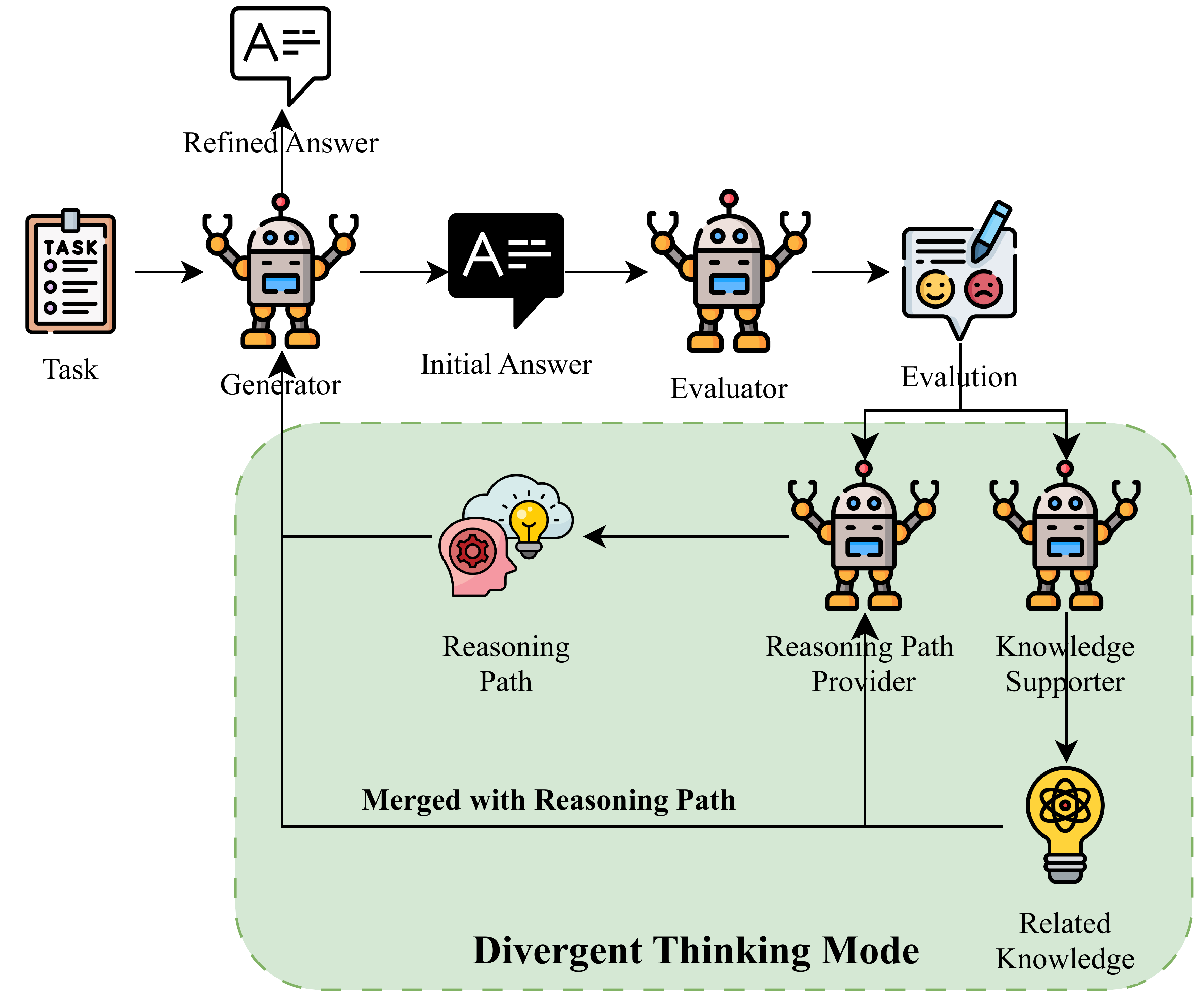}}
\caption{An Overview of Divergent Thinking Mode. Parallel proposals of knowledge and reasoning paths are synthesized into answers, benefiting commonsense tasks.}
\vspace{-8 mm}
\label{fig:divergent}
\end{center}
\end{figure}
\subsection{Divergent Thinking mode}
We present the framework of \textbf{Divergent Thinking Mode} in \cref{fig:divergent}
Within this framework, the \textbf{Divergent Thinking Mode} encompasses two agents except for the Generator and the Evaluator mentioned above. Each agent has a distinct role.\\

\textbf{Knowledge Supporter.} The Knowledge Supporter is primarily responsible for retrieving domain-specific knowledge relevant to the task, validating the accuracy and applicability of knowledge, and integrating multi-source knowledge to support answer generation. Its knowledge support forms the theoretical foundation of system output, ensuring the professionalism and reliability of the answers.

\textbf{Reasoning Path Provider.} The Reasoning Path Provider is responsible for constructing the reasoning path by designing optimal reasoning paths, validating the logical completeness of reasoning processes, and generating formalized reasoning paths. Its output reasoning path provides the system with a clear argumentative framework, ensuring the traceability and explainability of the answer generation. The reasoning path and related knowledge are jointly input into the Generator module. The Generator then modifies its initial answer based on these inputs, generating a refined answer through a systematic integration of reasoning paths and domain knowledge. This refined answer is subsequently channeled into the Discussion module. \\

Based on the above framework, we can construct the \textbf{Divergent Thinking Mode} with the specific formulas as follows:
\begin{equation}
    K=KnowledgeSupporter(E,A), 
\end{equation}
\begin{equation}
    R=PathProvider(E,K), 
\end{equation}
Where $E$ represents the evaluation draft, $A$ represents the initial answer, $K$ represents the related knowledge, $R$ represents reasoning paths. And we can have the refined answer as follows:
\begin{equation}
    O=Generator(R,K), 
\end{equation}
Where $O$ represents refined answer. Following the generation of the refined answer, we implement a discussion module that facilitates multi-agent deliberation to evaluate the correctness and soundness of the refined answer. The Discussion module, comprising Evaluator, Knowledge Supporter, and Reasoning Path Provider, serves as a collaborative debate verifier.\\

Within this module, these three agents engage in a structured debate regarding the validity of the refined answer. The outcome of this discussion follows a binary decision path in which if the refined answer meets the established criteria for correctness through consensus among the three agents, the debate process ends, and the answer is accepted as final; if discrepancies or inadequacies are identified in the refined answer, the system initiates another round of debate. This iterative process, depicted as "Next Round Debate" in the framework diagram, continues until either a satisfactory consensus is reached or the system reaches its predetermined maximum number of debate rounds.
\begin{figure}[t]
\begin{center}
\centerline{\includegraphics[width=\columnwidth]{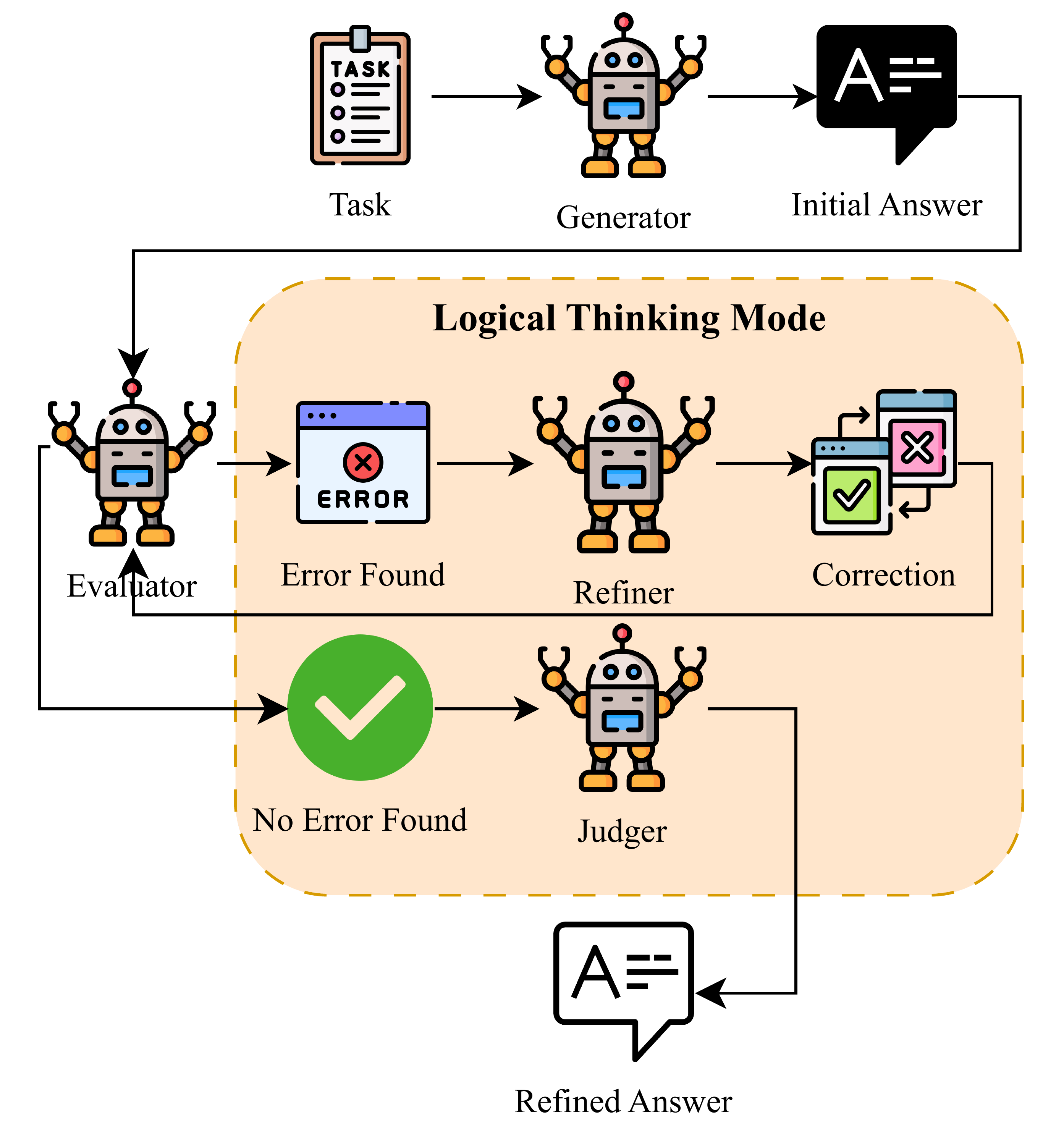}}
\caption{An Overview of Logical Thinking Mode. An evaluate–refine–judge loop enforces step‑wise verification, effective for mathematical reasoning.}
\vspace{-8 mm}
\label{fig:logical}
\end{center}
\end{figure}
\subsection{Logical Thinking Mode}
We provide an overview of the \textbf{Logical Thinking Mode} of DiMo in \cref{fig:logical}. In this mode, we also set up the Generator to receive the task inputs and generate the initial answer and the Evaluator to assess the initial answers alongside two agents tasked with debate.
\begin{table*}[t]
\caption{Dataset Statistics. Six benchmarks spanning commonsense and math reasoning (CSQA, OpenBookQA, ARC‑C, StrategyQA, GSM8K, GSM‑Hard), with dataset sizes and task characteristics summarized.}
\label{tab:dataset-stats}
\vskip 0.15in
\begin{center}
\begin{small}
\begin{sc}
\begin{tabular}{lccc}
\toprule
Dataset & Task Type & Task Info & Size \\
\midrule
CSQA\cite{talmor2018commonsenseqa} & Commonsense & 5 Choices & 1,221 \\
ARC-Challenge\cite{clark2018think} & Commonsense & 4 Choices & 1,170 \\
StrategyQA\cite{mihaylov2018can} & Commonsense & True/False & 687 \\
OpenBookQA\cite{geva2021did} & Commonsense & 4 Choices & 500 \\
GSM8K\cite{cobbe2021training} & Math & Questionnaire & 1,320 \\
GSM-hard\cite{gao2022pal} & Math & Questionnaire & 1,320 \\
\bottomrule
\end{tabular}
\end{sc}
\end{small}
\end{center}
\end{table*}
\begin{table*}[t]
\caption{The Overall Performance of DiMo and Baselines on Commonsense Reasoning Tasks. \textbf{Bold} numbers represent the performance of our method DiMo. \underline{Underlined} numbers represent the best performance.}
\label{tab:results}
\begin{center}
\begin{small}
\begin{sc}
\begin{tabular}{lcccc}
\toprule
Method & CSQA & ARC-Challenge & StrategyQA & Openbook QA \\
\midrule
LLaMA-3-8B & 76.1\% & 79.1\% & 63.4\% & 79.2\% \\
LLaMA-3-8B(CoT) & 72.5\% & 77.5\% & 74.2\% & 77.8\% \\
LLM MAD(LLaMA-3-8B) & 70.6\% & 78.3\% & 84.4\% & 74.3 \% \\
MAD(LLaMA-3-8B) & 46.2\% & 57.7\% & 42.3\% & 26.4\% \\ 
DiMo(LLaMA-3-8B) & \underline{\textbf{80.02\%}} & \underline{\textbf{84.1\%}} & \underline{\textbf{92.7\%}} & \underline{\textbf{84.5\%}} \\
\midrule
Qwen-2.5-32B & 87.6\% & 93.0\% & \underline{93.4\%} & 93.20\% \\
Qwen-2.5-32B(CoT) & 86.5\% & 94.0\% & 78.0\% & 95.40\% \\
QwQ-32B-preview & 80.0\% & 87.3\% & 81.3\% & 87.2\% \\
LLM MAD(Qwen-2.5-32B) & 86.3\% & \underline{94.1\%} & 91.4\% & 94.9\% \\
MAD(Qwen-2.5-32B) & 79.1\% & 91.4\% & 58.3\% & 23.4\% \\
DiMo(Qwen-2.5-32B) & \textbf{88.4\%} & \textbf{90.5\%} & \textbf{90.8\%} & \underline{\textbf{96.0\%}} \\
\bottomrule
\end{tabular}
\end{sc}
\end{small}
\end{center}
\end{table*}

\textbf{Refiner.} The Refiner is a correction agent that performs the refinement of problematic steps identified by the Evaluator. This agent employs a localized rewriting methodology, maintaining consistency with preceding and subsequent steps while implementing necessary corrections. The refinement corrects errors while preserving the reasoning path.\\

\textbf{Judger.} The Judger is a comprehensive assessment agent that conducts a holistic judgment of the refined reasoning path and final solution if it demonstrates both logical consistency and computational accuracy or initiates a single iteration of the Evaluator-Refiner cycle if deficiencies are detected. The judgment process is bounded by a predetermined iteration limit to prevent infinite recursion, ensuring computational efficiency while maintaining solution quality. \\

Based on the above framework, we can construct the \textbf{Logical Thinking Mode} with the specific formulas as follows:
\begin{equation}
    e=Evaluator(A), 
\end{equation}
Where $e$ represents the error status, $A$ represents the initial answer. Then we can have conditions:
\begin{equation}
    R = \begin{cases}
Refiner(A) & \text{if } e = 1 \\
Judger(A) & \text{if } e = 0
\end{cases},
\end{equation}
Where $R$ represent the refined answer. Finally, the entire mode can be formalized through the following formula:
\begin{equation}
    R = Evaluator(A)^n \cdot Refiner^n(A) + (1-Evaluator(A)^n) \cdot Judger(A), 
\end{equation}
Where $n$ is a number of Evaluator-Refiner correction cycles.

\section{Experiment}
\label{sec}

In our experiment, we evaluate our multi-agent debate system using different types of QA tasks, which test whether our multi-agent debate system improves LLM reasoning.

\subsection{Tasks \& Datasets}
Our experiments assess knowledge-based and math tasks.\\
\textbf{Tasks}. Our experiments use 4 commonsense/knowledge-based and 2 math reasoning datasets.\\

\textbf{Commonsense.} For these tasks, we focus on evaluating whether LLMs could enhance their understanding of implicit knowledge within problems and generate appropriate reasoning paths through collaborative discussions between models, thus improving the accuracy of their final solutions.\\

\textbf{Math.} For these tasks, we evaluate whether LLMs could rectify mathematical reasoning errors through multi-agent collaboration while generating correct reasoning paths, thus improving their accuracy in mathematical reasoning tasks.\\

All datasets are listed in \cref{tab:dataset-stats}.
We choose CommonsenseQA \cite{talmor2018commonsenseqa}, ARC-Challenge \cite{clark2018think}, OpenBookQA \cite{geva2021did}, and StrategyQA \cite{mihaylov2018can} for the commonsense and knowledge-based reasoning task. And we choose GSM8K \cite{cobbe2021training} and GSM-hard \cite{gao2022pal} for the math reasoning task.

\subsection{Baselines} We define four types of baselines in our experiment.\\

\textbf{Large Language Models.} We select two open-source base models: LLaMA-3-8B \cite{llama3modelcard} and Qwen-2.5-32B \cite{yang2024qwen2} as the backbone for all agents. This pair spans distinct families and parameter scales ($\approx$8B vs. 32B), enabling compute‑controlled comparisons in a reproducible and license‑compatible setup. To ensure fairness, the same checkpoint is shared across roles; only system prompts and decoding temperatures differ between modes (divergent uses higher temperature; logical uses lower). Unless otherwise noted, we keep context window, sampling strategy, and per‑item token budgets identical across baselines. Our analysis standardizes the backbone to LLaMA‑3‑8B and Qwen‑2.5‑32B, centering the study on protocol differences rather than cross‑family variation.\\

\textbf{Chain-of-Thought Prompting.} We include a single‑model baseline that elicits step‑wise rationales via a neutral instruction and then extracts a short final answer. Previous works \cite{wei2022chain,zhang2022automatic} demonstrates CoT enhances LLM reasoning capabilities. To ensure fair comparison, we use the same backbone, sampling strategy, and per‑item token budgets as other baselines; rationale tokens are counted toward the budget, and answer extraction follows a deterministic parser for multiple‑choice and numeric outputs.\\

\textbf{o1-like Reasoning LLMs.} Since the release of OpenAI's GPT-o1 model, this post-training method combining reinforcement learning and Chain-of-Thought has been demonstrated to effectively enhance LLMs' performance on mathematical reasoning tasks. We select QwQ-32B-preview\cite{qwq-32b-preview}, an open-source o1-like model fine-tuned on Qwen-2.5-32B, as the baseline model. For fairness, decoding settings and per‑item token budgets match those of other baselines, and all rationale tokens are included in token accounting. Tool use or external code execution is disabled to isolate the effect of the model’s internal reasoning from external resources.\\

\textbf{Multi Agent Debate.} We select two latest multi-agent debate methods to serve as baselines in our experiments: one is the LLM MAD(LLM Multi-agent Debate) method proposed by Du et al.,\cite{du2023improving}, the other is the MAD method introduced by Liang et al.\cite{liang2023encouraging} For compute parity, we freeze the backbone across roles, align decoding parameters with other baselines, and enforce identical per‑item token budgets; all debate messages count toward token accounting. 
\subsection{Evaluation Metrics} 
Currently, the majority of existing datasets utilize Exact Match(EM)\cite{jiang2023active}. In our experimental design, we implement prompt engineering to ensure LLMs output strictly conforms to dataset-specific answer formats. The selected datasets predominantly feature multiple-choice, binary judgment, and numerical filled-in-the-blanks. Given these output requirements, we adopt Exact Match as evaluation metrics. For multiple‑choice we map option texts/letters to a canonical label; for yes/no we normalize to “yes”/“no”; and for numeric answers we strip whitespace, punctuation, and unit variants. 
\subsection{Experiments Settings}
\textbf{LLMs Used in the Agents} The experimental implementation utilized consistent model architecture across both our agents and the baselines, specifically employing LLaMA-3-8B and Qwen-2.5-32B. 
\begin{table}[t]
\caption{The Overall Performance of DiMo and Baselines on Mathematical Reasoning Tasks. \textbf{Bold} numbers represent the performance of our method DiMo. \underline{Underlined} numbers represent the best performance.}
\label{tab:math-results}
\begin{center}
\begin{small}
\begin{sc}
\begin{tabular}{lcc}
\toprule
Method                & GSM8K                                     & GSM-hard                                  \\
\midrule
LLaMA-3-8B            & 49.8\%                                   & 42.3\%                                   \\
LLaMA-3-8B(CoT)       & 84.0\%                                   & 44.7\%                                   \\
LLM MAD(LLaMA-3-8B)   & 75.8\%                                   & 34.6\%                                    \\
MAD(LLaMA-3-8B)       & 43.7\%                                   & 15.6\%                                    \\
DiMo(LLaMA-3-8B)    & \textbf{\underline{90.7\%}} & \textbf{\underline{71.4\%}} \\
\midrule
Qwen-2.5-32B          & 37.3\%                                   & 28.9\%                                   \\
Qwen-2.5-32B(CoT)     & 95.2\%                                   & 72.4\%                                   \\
QwQ-32B-preview       & 75.4\%                                   & 51.4\%                                   \\
LLM MAD(Qwen-2.5-32B) & 94.7\%                                   & 60.9\%                                    \\
MAD(Qwen-2.5-32B)     & 89.8\%                                   & 30.9\%                                    \\
DiMo(Qwen-2.5-32B)  & \textbf{\underline{98.4\%}} & \textbf{\underline{84.1\%}} \\ 
\bottomrule
\end{tabular}
\end{sc}
\end{small}
\end{center}
\end{table}
\begin{table}[t]
\caption{The Results of Using Different Modes on Mathematical Reasoning Tasks. Logical mode outperforms Divergent on math, supporting a protocol–task affinity for step‑wise reasoning.}
\label{tab:modes-comparison}
\begin{center}
\begin{small}
\begin{sc}
\begin{tabular}{lcc}
\toprule
Method & GSM8k & GSM-hard \\
\midrule
DiMo & & \\
(Divergent Thinking Mode) & 79.9\% & 57.2\% \\
(LLaMA-3-8B) & & \\
\midrule
DiMo & & \\
(Logical Thinking Mode) & 90.7\% & 71.4\% \\
(LLaMA-3-8B) & & \\
\midrule
LLaMA-3-8B(CoT) & 84.0\% & 44.7\% \\
\bottomrule
\end{tabular}
\end{sc}
\end{small}
\end{center}
\end{table}
\subsection{Main Results}
\textbf{Commonsense Reasoning Tasks.} For these tasks, DiMo will use \textbf{Divergent Thinking Mode}. In \cref{tab:results}, we present the results of our method on four commonsense and knowledge-based reasoning tasks. From \cref{tab:results}, DiMo outperforms baselines in most knowledge-based datasets. Our experiment finds that the results reveal significant performance improvements on multiple-choice tasks(CSQA, ARC-Challenge, and OpenBookQA) using our DiMo method. While most of the baselines using LLaMA-3-8B achieve accuracy below 80\% on all three tasks, DiMo using LLaMA-3-8B consistently elevates the accuracy above 80\%. Furthermore, DiMo achieves higher accuracy on binary(True/False) judgment tasks like StrategyQA. While the baseline LLM MAD achieves 84.4\%, DiMo utilizing LLaMA-3-8B achieves a higher accuracy of 92.7\%.\\

\textbf{Mathematical Reasoning Tasks.} For these tasks, DiMo will use \textbf{Logical Thinking Mode}. In \cref{tab:math-results}, DiMo demonstrates superior performance compared to the baselines when using both LLaMA-3-8B and Qwen-2.5-32B. To explore the potential of DiMo in enhancing LLMs' mathematical reasoning capabilities, we utilized the GSM-hard dataset, which increased the difficulty of GSM8K benchmark. Our approach significantly improves performance on challenging mathematical reasoning tasks, as shown in \cref{tab:math-results}\\

\textbf{Comparison with o1-like Reasoning LLMs.} DiMo demonstrates superior performance not only on general LLMs but also when evaluated against Large Reasoning Language Models that employ post-training techniques, including reinforcement learning and Chain-of-Thought prompting. As shown in \cref{tab:results} and \cref{tab:math-results}, DiMo utilizing Qwen-2.5-32B consistently outperforms the QwQ-32B-preview in both commonsense reasoning and mathematical reasoning tasks, demonstrating significant accuracy improvements.

\section{Discussion}
\subsection{Different Tasks Require Different Thinking Modes}
Our experiments reveal that different reasoning tasks are optimally processed by Large Language Models(LLMs) through distinct thinking modes. This aligns with human cognitive patterns, where factual knowledge-based problems typically engage associative memory and divergent exploration of knowledge points to arrive at solutions. In contrast, when confronting logical reasoning problems, humans tend to employ systematic, step-by-step deductive processes. 

In our experiment, we use the \textbf{Divergent Thinking Mode} to process mathematical reasoning tasks. The results are presented in \cref{tab:modes-comparison}.
As demonstrated in \cref{tab:modes-comparison}, DiMo exhibits lower accuracy when employing the \textbf{Divergent Thinking Mode} for GSM8k and GSM-hard datasets compared to using the \textbf{Logical Thinking Mode}. Notably, its performance on the GSM8k dataset even falls below the accuracy achieved by LLaMA-3-8B with Chain-of-Thought prompting. These findings confirm that LLMs, like humans, need task-specific thinking modes to optimize problem-solving.

\subsection{Different Initialization Prompts}
In our experiment, we observed that when DiMo processes commonsense reasoning tasks, variations in initial prompts demonstrate a measurable impact on the final accuracy.
\begin{table}[t]
\caption{The Accuracy of DiMo on CSQA by Using Different Prompts. No‑CoT initialization exceeds CoT on CSQA, highlighting prompt design’s impact on convergence and accuracy.}
\label{tab:prompt-comparison}
\begin{center}
\begin{small}
\begin{sc}
\begin{tabular}{lcc}
\toprule
Method & Prompts & CSQA \\
\midrule
DiMo & CoT & 75.81\% \\
(LLaMA-3-8B) & No CoT & 80.02\% \\
\bottomrule
\vspace{-8 mm}
\end{tabular}
\end{sc}
\end{small}
\end{center}
\end{table}
We conduct two experiments with distinct initial prompts: one group incorporates Chain-of-Thought(CoT) prompts in the Generator's initial prompts, while the other group operates without CoT prompts, allowing the Generator to output the initial answer directly. The results of the experiment are presented in \cref{tab:prompt-comparison}. We find that using CoT prompts in the Generator's initial prompts, rather than making the Generator output the answer directly, leads to a decrease in the final accuracy on the CSQA dataset. This decline may be related to the reasoning process introduced by the CoT prompts. CoT prompts are designed to guide the model through step-by-step reasoning, providing more intermediate steps, which could increase the complexity. Despite sometimes improving reasoning in other contexts, CoT prompts in this experiment decreased accuracy by introducing interference or causing over-reliance on initial reasoning steps.
\begin{figure}[h]
\begin{center}
\centerline{\includegraphics[width=\columnwidth]{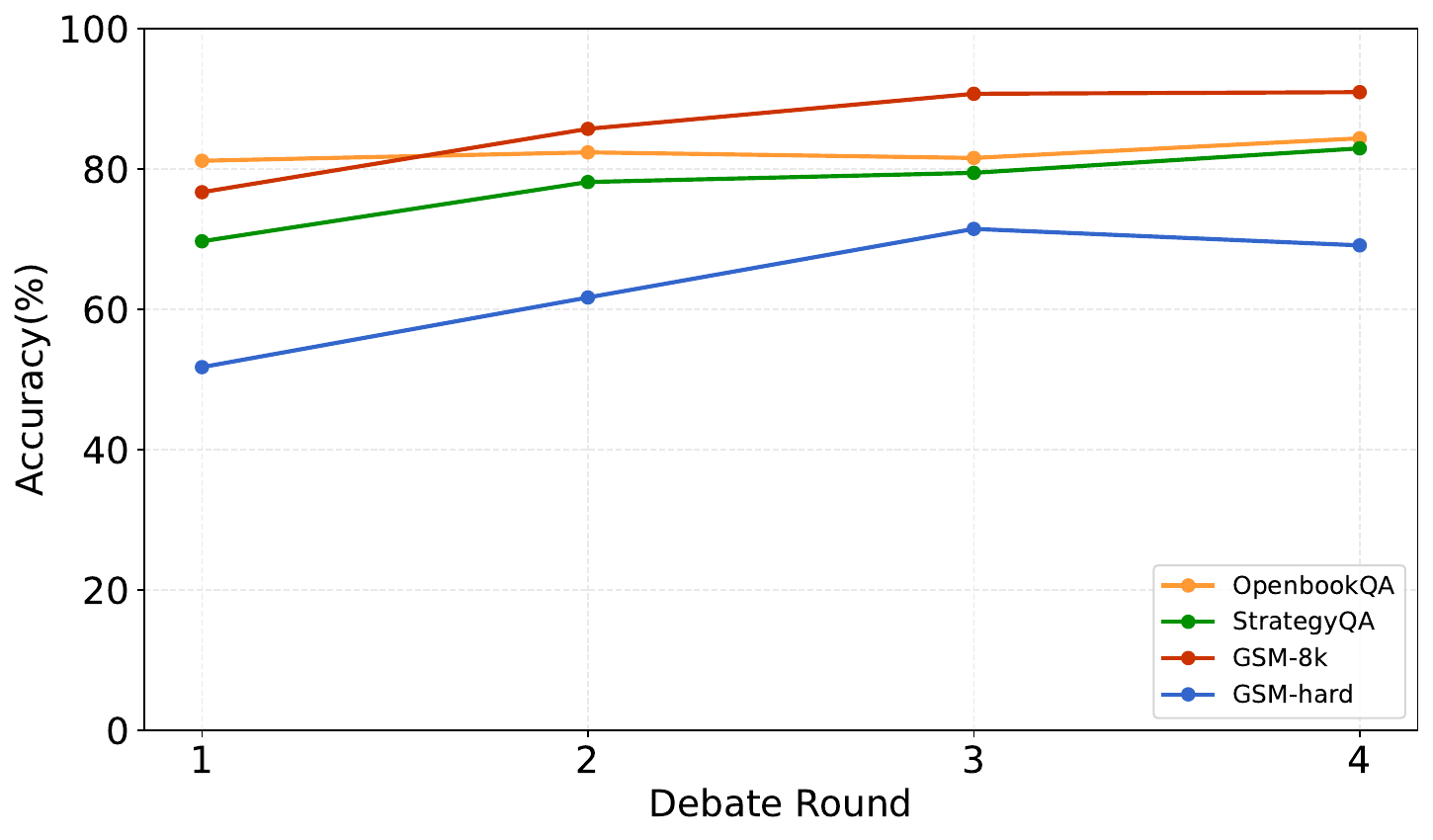}}
\caption{Accuracy of DiMo with Different Debate Rounds. Accuracy improves up to about three rounds, and the subsequent upward trend has leveled off.}
\vspace{-8 mm}
\label{fig:rounds}
\end{center}
\end{figure}
\subsection{Number of Debate Rounds}
To explore the impact of varying debate rounds between multiple agents on the final accuracy among different tasks, we design four groups of experiments with different debate rounds. The results are presented in \cref{fig:rounds}. 
Our experimental design incorporates four rounds of debate. As illustrated in \cref{fig:rounds}, the OpenBookQA dataset and StrategyQA dataset exhibit relatively stable accuracy rates when the number of debate rounds exceeds one, fluctuating around 80\%. In contrast, the GSM8K and GSM-hard dataset demonstrate a positive correlation between accuracy and debate rounds, with performance improving substantially from 76\% to 90\% and 50\% to approximately 70\% as the number of rounds increases. And we can find that the accuracy rates for both the GSM8K and GSM-Hard datasets peaked at 3 rounds of debate.
\subsection{Tasks Token Counting for Computational Consumption}
We present the token counts for the CommonsenseQA task and the GSM8K task in \cref{tab:token-counts}.The comparative analysis reveals that while our multi-agent debate framework incurs elevated token consumption on CommonsenseQA and GSM8K benchmarks relative to singular-model approaches, the strategic implementation of cost-effective open-source LLMs (e.g., LLaMA/Qwen) renders this differential economically marginal. Rather than optimizing for incremental performance, we advance interpretability with cognition-inspired architectures that combine task-specific reasoning simulation with adversarial hypothesis refinement, enabling reproducible, cross-task analyses of LLM reasoning processes and failure modes.
\begin{table}[h]
\caption{Token Counts for Different Tasks. Token counts reveal an accuracy–cost trade‑off: multi‑agent methods use more tokens yet yield auditable traces.}
\label{tab:token-counts}
\begin{center}
\begin{small}
\begin{sc}
\begin{tabular}{lcc}
\toprule
Models & CSQA & GSM8K  \\
\midrule
LLaMA-3-8B & 95 & 106 \\
LLaMA-3-8B(CoT) & 430 & 369 \\
DiMo(LLaMA-3-8B) & 19300 & 1929 \\
Qwen-2.5-32B & 100 & 105 \\
Qwen-2.5-32B(CoT) & 500 & 410 \\
DiMo(Qwen-2.5-32B) & 13700 & 2650 \\
\bottomrule
\vspace{-8 mm}
\end{tabular}
\end{sc}
\end{small}
\end{center}
\end{table}
\begin{figure}[h]
\begin{center}
\centerline{\includegraphics[width=0.95\columnwidth]{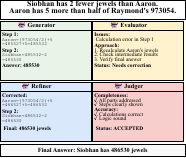}}
\caption{Case study of DiMo on GSM-hard dataset. The Evaluator flags errors, the Refiner applies targeted fixes, and the Judge finalizes a transparent solution.}
\vspace{-8 mm}
\label{fig:case}
\end{center}
\end{figure}
\subsection{Case Study}
To analyze the performance of agents during the debate and collaboration process, we randomly select a case from the GSM-hard dataset for analysis, which is presented in \cref{fig:case}. As illustrated in \cref{fig:case}, the process demonstrates effective error detection and correction through multiple stages. After the Generator outputs the initial answer, the Evaluator receives the initial answer and identifies a computational error in Step 1 of the Generator's initial answer. Upon receiving this feedback, the Refiner successfully recognizes and corrects the error, ultimately producing an accurate solution that passes the Judger's validation process. This case effectively illustrates how DiMo enhances LLMs' reasoning capabilities through collaborative multi-agent debate. The framework demonstrates its effectiveness in three key areas: error detection through inter-agent cooperation, systematic error correction, and the generation of accurate final answers. Furthermore, DiMo produces explicit reasoning paths, substantially improving LLMs' interpretability. Throughout, “interpretability” refers to process transparency—explicit intermediate states that can be checked by other agents or humans. We do not make claims about the mechanistic interpretability of base models. The Case study illustrates error localization and evidence auditing; establishing task‑agnostic quantitative measures is the future work.
\section{Conclusion}

In this paper, we address two critical challenges for Large Language Models (LLMs): first, their varying preferences for distinct thinking modes across different categories of reasoning tasks; and second, the limited interpretability of their internal reasoning processes. To tackle these issues in a unified manner, we propose the Multi-Agent Collaboration Framework for Diverse Thinking Modes (DiMo), which explicitly incorporates two complementary modes of thought: a Logical Thinking Mode, specifically designed for mathematical reasoning tasks that benefit from structured deduction, and a Divergent Thinking Mode, designed for commonsense reasoning tasks that require breadth, flexibility, and associative exploration. Across the majority of evaluated datasets, DiMo consistently outperforms baseline methods; notably, on the more challenging GSM-hard dataset, our approach achieves an accuracy that exceeds the baseline by over 20\%, underscoring a substantial improvement in the reasoning capabilities of LLMs. Beyond raw performance, DiMo also generates explicit, human-auditable reasoning paths, thereby significantly enhancing the interpretability and transparency of the overall reasoning process. Taken together, these results validate the effectiveness of DiMo both in strengthening LLM reasoning and in providing clearer insight into how solutions are derived. Looking ahead, we will extend evaluation to more challenging mathematical and scientific reasoning datasets with varied answer formats—such as MATH \cite{saxton2019mathematical} and GPQA \cite{rein2023gpqa}—as well as to additional forms of reasoning, including symbolic reasoning tasks.

\bibliographystyle{ACM-Reference-Format}
\bibliography{custom}

\appendix
\onecolumn
\section{Role Prompts}

\begin{table}[h]
\caption{The Role Descriptions and Prompts for Different Agents in Divergent Thinking Mode of MA4DTM}
\label{tab:prompts}
\vskip 0.15in
\begin{center}
\begin{small}
\begin{tabular}{ll}
\toprule
Role & Prompt \\
\midrule
Evaluator & You are an answer evaluator for multiple-choice \\
& commonsense reasoning questions. Analyze the \\
& given answer for: \\
& 1. Logical reasoning \\
& 2. Consideration of all options \\
& 3. Understanding of common sense principles \\
& 4. Clear justification for the chosen answer \\
& \\
& Please point out any errors, omissions, or areas \\
& that need improvement. Focus on whether the \\
& reasoning is sound and leads to a logical conclusion. \\
\midrule
Knowledge & You are a knowledge supporter for multiple-choice \\
Supporter & commonsense reasoning questions. Your role is to: \\
& 1. Identify relevant commonsense knowledge needed \\
& for the question \\
& 2. Provide real-world examples and context \\
& 3. Highlight any missing or incorrect information \\
& 4. Support or challenge the reasoning with factual \\
& information \\
& \\
& Focus on practical, everyday knowledge that relates \\
& to the question and choices. \\
\midrule
Reasoning & You are a reasoning path provider for multiple-choice \\
Path & commonsense reasoning questions. Create clear \\
Provider & logical paths that: \\
& 1. Break down the question into key components \\
& 2. Analyze each option systematically \\
& 3. Show clear steps leading to the answer \\
& 4. Explain why incorrect options are eliminated \\
& 5. Connect evidence to conclusions \\
& \\
& Ensure each step follows logically from the previous \\
& one and leads to a clear choice. \\
\bottomrule
\end{tabular}
\end{small}
\end{center}
\end{table}

\cref{tab:prompts} presents the role sample prompts for agents in the Divergent Thinking mode of MA4DTM. These sample prompts are used for the CommonsenseQA dataset.

\newpage
\onecolumn
\section{Dataset Format}

\subsection{Commonsense Reasoning Datasets Details}

\begin{table}[h]
\caption{CommonsenseQA Sample Format}
\label{tab:question-analysis}
\vskip 0.15in
\begin{center}
\begin{small}
\begin{tabular}{lcl}
\toprule
Components & & Content \\
\midrule
Question & & A revolving door is convenient for two direction travel, \\
& & but it also serves as a security measure at a what? \\
Question Concept & & revolving door \\
Choices & & \{ "label": [ "A", "B", "C", "D", "E" ], \\
& & "text": [ "bank", "library", "department store", \\
& & "mall", "new york" ] \} \\
Answer Key & & A \\
\bottomrule
\end{tabular}
\end{small}
\end{center}
\vskip -0.1in
\end{table}

\begin{table}[h]
\caption{ARC-Challenge Sample Format}
\label{tab:astronomy-question}
\vskip 0.15in
\begin{center}
\begin{small}
\begin{tabular}{lcl}
\toprule
Components & & Content \\
\midrule
Question & & An astronomer observes that a planet rotates faster after \\
& & a meteorite impact. Which is the most likely effect of this \\
& & increase in rotation? \\
Choices & & \{ "text": [ "Planetary density will decrease.", \\
& & "Planetary years will become longer.", \\
& & "Planetary days will become shorter.", \\
& & "Planetary gravity will become stronger." ], \\
& & "label": [ "A", "B", "C", "D" ] \} \\
Answer Key & & C \\
\bottomrule
\end{tabular}
\end{small}
\end{center}
\vskip -0.1in
\end{table}

\begin{table}[h]
\caption{OpenbookQA Sample Format}
\label{tab:digestion-question}
\vskip 0.15in
\begin{center}
\begin{small}
\begin{tabular}{lcl}
\toprule
Components & & Content \\
\midrule
Question & & When food is reduced in the stomach \\
Choices & & \{ "text": [ "the mind needs time to digest", \\
& & "take a second to digest what I said", \\
& & "nutrients are being deconstructed", \\
& & "reader's digest is a body of works" ], \\
& & "label": [ "A", "B", "C", "D" ] \} \\
Answer Key & & C \\
Fact & & digestion is when stomach acid breaks down food \\
Humanscore & & 1 \\
Clarity & & 1.6 \\
\bottomrule
\end{tabular}
\end{small}
\end{center}
\vskip -0.1in
\end{table}

\newpage
\onecolumn

\begin{table}[h]
\caption{StrategyQA Sample Format}
\label{tab:strategy-question}
\vskip 0.15in
\begin{center}
\begin{small}
\begin{tabular}{lcl}
\toprule
Components & & Content \\
\midrule
Question & & Was ship that recovered Apollo 13 named after a \\
& & World War II battle? \\
Answer & & True \\
Term & & Apollo 13 \\
Description & & A failed crewed mission to land on the Moon \\
Facts & & Apollo 13 was recovered by the USS Iwo Jima. \\
& & Iwo Jima was captured from the Imperial Japanese \\
& & Army during World War II by the US in a conflict \\
& & called the Battle of Iwo Jima. \\
\bottomrule
\end{tabular}
\end{small}
\end{center}
\vskip -0.1in
\end{table}

\cref{tab:question-analysis,tab:astronomy-question,tab:digestion-question,tab:strategy-question} present the commonsense reasoning datasets sample that used in our experiments. Our datasets consist of three multiple-choice dataset and one binary judgment dataset. 

\subsection{Mathematical Reasoning Datasets Details}

\begin{table}[h]
\caption{GSM8K Sample Format}
\label{tab:word-problem}
\vskip 0.15in
\begin{center}
\begin{small}
\begin{tabular}{lcl}
\toprule
Components & & Content \\
\midrule
Question & & Janet's ducks lay 16 eggs per day. She eats three for \\
& & breakfast every morning and bakes muffins for her \\
& & friends every day with four. She sells the remainder \\
& & at the farmers' market daily for \$2 per fresh duck \\
& & egg. How much in dollars does she make every day \\
& & at the farmers' market? \\
Answer & & Janet sells 16 - 3 - 4 = <<16-3-4=9>>9 duck eggs \\
& & a day. She makes 9 * 2 = \$<<9*2=18>>18 every day \\
& & at the farmer's market. \#\#\#\# 18 \\
\bottomrule
\end{tabular}
\end{small}
\end{center}
\vskip -0.1in
\end{table}

\begin{table}[h]
\caption{GSM-hard Sample Format}
\label{tab:chicken-feed}
\vskip 0.15in
\begin{center}
\begin{small}
\begin{tabular}{lcl}
\toprule
Components & & Content \\
\midrule
Input & & Every day, Wendi feeds each of her chickens three cups \\
& & of mixed chicken feed, containing seeds, mealworms and \\
& & vegetables to help keep them healthy. She gives the \\
& & chickens their feed in three separate meals. In the \\
& & morning, she gives her flock of chickens 6887483 cups \\
& & of feed. In the afternoon, she gives her chickens \\
& & another 25 cups of feed. How many cups of feed does \\
& & she need to give her chickens in the final meal of \\
& & the day if the size of Wendi's flock is 20 chickens? \\
Target & & -6,887,448 \\
\bottomrule
\end{tabular}
\end{small}
\end{center}
\vskip -0.1in
\end{table}
\cref{tab:word-problem,tab:chicken-feed} present the mathematical reasoning datasets sample used in our experiments.

\end{document}